\documentclass[conference]{IEEEtran}
\usepackage[T1]{fontenc}
\IEEEoverridecommandlockouts
\usepackage{cite}
\usepackage{textcomp}
\usepackage{xcolor}
\def\BibTeX{{\rm B\kern-.05em{\sc i\kern-.025em b}\kern-.08em
    T\kern-.1667em\lower.7ex\hbox{E}\kern-.125emX}}
\usepackage{amsmath,graphicx,tikz,amssymb,amsfonts,bm,mathtools,mathabx}
\usepackage{multicol}
\usepackage{amsfonts}
\usepackage{amssymb}
\usepackage{mathrsfs}
\usepackage{amsthm}
\usepackage{float}

\usepackage{comment}
\usepackage[linkcolor=mar,colorlinks=true]{hyperref}
\hypersetup{colorlinks,breaklinks,
            linkcolor=[rgb]{0.5,0,0},
            citecolor=[rgb]{0.39,0.7,1.0}}
\usepackage{cleveref}
\usepackage{cite}
\usepackage{enumitem}
\usepackage{outlines}
\usepackage[ruled,vlined,linesnumbered]{algorithm2e}
\usepackage{algpseudocode}
\usepackage[skip=0pt]{caption}
\usepackage[skip=0pt]{subcaption}
\usepackage{bm}
\newtheorem{theorem}{Theorem}
\colorlet{lightgray}{blue!5.5}
\colorlet{orange}{orange!10}
\newtheoremstyle{reduced}
  {6pt} 
  {6pt} 
  {} 
  {} 
  {\bfseries} 
  {.} 
  {.5em} 
  {} 

\theoremstyle{reduced}
\newtheorem{assumption}{Assumption}
\newtheorem{remark}{Remark}

\begin{document}


\title{Strong Duality Relations in Nonconvex\\\hspace{5.5pt}Risk-Constrained Learning
\thanks{The authors would like to kindly acknowledge support by a Microsoft gift, as well as by the US National Science Foundation (NSF) under Grant CCF 2242215. Author names are listed in alphabetical order.}}

\author{\IEEEauthorblockN{Dionysis Kalogerias}
\IEEEauthorblockA{\textit{Department of Electrical Engineering} \\
\textit{Yale University}\\
New Haven, USA \\
dionysis.kalogerias@yale.edu}
\and
\IEEEauthorblockN{Spyridon Pougkakiotis}
\IEEEauthorblockA{\textit{School of Science and Engineering} \\
\textit{University of Dundee}\\
Dundee, UK \\
spougkakiotis001@dundee.ac.uk}}

\maketitle
\begin{abstract}
We establish strong duality relations for functional two-step compositional risk-constrained learning problems with multiple nonconvex loss functions and/or learning constraints, regardless of nonconvexity and under a minimal set of technical assumptions. Our results in particular imply zero duality gaps within the class of problems under study, both extending and improving on the state of the art in (risk-neutral) constrained learning. More specifically, we consider risk objectives/constraints which involve real-valued convex and positively homogeneous risk measures admitting dual representations with bounded risk envelopes, generalizing expectations and including popular examples, such as the conditional value-at-risk (CVaR), the mean-absolute deviation (MAD), and more generally all real-valued coherent risk measures on integrable losses as special cases. Our results are based on recent advances in risk-constrained nonconvex programming in infinite dimensions, which rely on a remarkable new application of J. J. Uhl’s convexity theorem, which is an extension of A. A. Lyapunov’s convexity theorem for general, infinite dimensional Banach spaces. By specializing to the risk-neutral setting, we demonstrate, for the first time, that constrained classification and regression can be treated under a unifying lens, while dispensing certain restrictive assumptions enforced in the current literature, yielding a new state-of-the-art strong duality framework for nonconvex constrained learning.
\end{abstract}
\vspace{2pt}

\vspace{-8pt}
\section{Introduction}
\vspace{-2pt}
\par Classification and regression tasks constitute two core problem classes widely appearing in numerous flavors in signal processing, machine and statistical learning. Recent advances in machine learning and artificial intelligence have rejuvenated this area of research, enabling highly efficient solution methods for wide classes of regression and classification tasks of paramount importance in practical applications, including, among many others, healthcare (e.g., \cite{DeepLearning_NatureMed}), engineering (e.g., \cite{ML_Applications_eng,ProcIEEE_SmartSystems}) and computer science (e.g., \cite{ML_Appl_CS}). Nonetheless, the literature on classification and regression has, so far, mostly focused on the unconstrained setting, often leading to standard fitting problems. While this class of problems is highly important, it fails to ensure that the associated learnt policy will explicitly satisfy key properties of interest, such as safety or lack of bias (see, e.g., \cite{NeuralNet_Robustness,Bias_in_ML1}). In such cases, highly specialized priors have been utilized, requiring extensive tuning and often unrealistic assumptions, thus leading to fragile methodologies (e.g. see \cite{pmlr-v48-cohenc16}). In light of the ubiquitous nature of systems utilizing regression or classification tasks, there has been a rise in the demand of appropriate ways to incorporate constraints for handling issues like fairness (e.g., \cite{pmlr-v65-woodworth17a,NIPS2016_dc4c44f6}), robustness (e.g., \cite{iclr_adversarial_learning,cianfarani2022understanding}), or safety (e.g., \cite{pmlr-v70-achiam17a,Paternain_etal_CDC}), among others. This is especially important in the context of nonconvex (constrained) learning, which typically occurs when utilizing (deep) neural networks (NNs).
\par Standard machine learning literature, which focuses on the unconstrained case, relies on the use of appropriate penalty functions to satisfy the various specifications that a given policy needs to abide by (e.g. \cite{pmlr-v80-xu18h}). However, designing appropriate penalties and finding the corresponding parameters for weighting the different objectives is a notorious problem, leading to practically questionable results or fragile ``optimal" policies. Such difficulties have recently lead to the study of constrained learning problems (e.g., \cite{JOTA_Shapiro,primal_dual_constrained_learning,NEURIPS2021_Chamon,Chamon2021}).
\par In this setting, one key question concerns the relation of the constrained learning problem with its associated \textit{Lagrangian dual}. This is particularly important to study in the context of nonconvex constrained learning, in which duality is an especially complicated concept. Specifically, \emph{strong duality} results ---which indicate that the primal problem can be substituted by the dual using certain Lagrange multipliers--- in constrained learning with general nonconvex losses in both the objective and/or the constraints are particularly important. Indeed, typical constrained learning problems and associated solution methods rely on some form of NN parametrization, which usually leads to nonconvex optimization problems (thus nonconvex constrained learning problems are ubiquitous). At the same time, parametrization in the primal domain allows one to tackle the problem in the dual domain, which is typically finite-dimensional (despite the infinite-dimensional nature of the original (non-parametrized) primal problem).

\par It is thus not a surprise that constrained learning problems are typically tackled in the Lagrangian dual domain (e.g. see \cite{pmlrzhang19p,primal_dual_constrained_learning,NEURIPS2021_Chamon,HounieRibeiroChamon,Paternain_etal_CDC}). Consequently, it is of key importance to find minimal conditions under which the problem under consideration exhibits strong duality, and especially in cases where the associated loss functions are nonconvex. Of course, in the context of convex learning this is relatively straightforward via utilizing standard constraint qualifications (CQ), such as Slater's CQ. In the nonconvex setting, however, this question is substantially more challenging. At the same time, strong duality in this context allows for the general study of generalization properties of the associated sample-average approximation (SAA) ---or empirical risk minimization (ERM)--- problems (as in \cite{JOTA_Shapiro,pmlr-v65-woodworth17a,NIPS2016_dc4c44f6,Chamon2021}), generalizing previous established results in the context of traditional statistical learning theory (e.g. see \cite{Vapnikbook}). Except for the above, strong duality is also important for establishing minimal conditions under which saddle points exist for the primal-dual problem, as well as for deriving necessary conditions for optimality. 
\par To the best of our knowledge, current state-of-the-art results concerning strong-duality in nonconvex constrained learning were recently developed in \cite{Chamon2021}. The theory in \cite{Chamon2021} distinguishes between two cases: constrained classification and constrained regression. In each case, by devising appropriate assumptions, the authors are able to establish strong duality relations in the general nonconvex setting, via utilizing the celebrated convexity theorem of A. A. Lyapunov (see \cite[Corollary IX.1.6]{Diestel1977}). While the results of \cite{Chamon2021} are fairly general, there are several issues which we believe can and should be addressed, strengthening the theory of  \cite{Chamon2021} significantly.
\par First, treating the classification and the regression tasks separately introduces some counter-intuitive and hard-to-evaluate assumptions (e.g., see \cite[Assumption 6]{Chamon2021}). Instead, our claim herein is that the classification and regression tasks can be naturally studied in a unified manner. Second, \cite{Chamon2021} assumes that the underlying functional space of feasible policies is closed, decomposable, and convex. We claim that only decomposability is required (as far as strong duality is concerned), and this is quite important, since convexity is often a property that decomposable spaces fail to satisfy (except in trivial cases). Third, we address some technical issues in \cite[Proof of Proposition III.2]{Chamon2021}, and propose a slightly different approach to resolve them. Apart from the above, our analysis also relaxes some additional assumptions utilized in \cite{Chamon2021}. 

In a nutshell, our contributions in this work enable the use of \textit{nonlinear risk functionals} in place of linear operators in the underlying model, i.e., our technical approach allows for general \textit{risk measures} \cite{Shapiro_Stochastic_Prog} in place of (linear) expectations. In fact, risk-aware (constrained) learning is an increasingly important subject that has recently gained significant traction (e.g. see \cite{JMLR:v19:17-295,DRlearning_Foundations,Duchi_DRO_learning}). There are several reasons for this. For instance, risk-aware learning can be incorporated to boost the tail behaviour of a classifier (e.g., see \cite{NEURIPS2021_b691334c}). Additionally, while on-average behaviour of existing systems might be optimal, they face failure when faced with atypical data (see \cite{iclr_dro_learning}). Further, by utilizing risk measures such as the CVaR, one can substantially robustify indicator-type (chance) constraints that are typically employed in the constrained learning literature (e.g., as in \cite[Section V.B]{Chamon2021}), thus resulting in more favorable problem formulations. Overall, incorporating risk-measures in constrained learning has a plethora of useful applications. 

Nonetheless, the focus of this work is mostly theoretical; indeed, we lay the theoretical foundations on how one can incorporate such risk measures in the objective and/or the constraints of a learning problem, and establish strong duality relations of such problems under minimal conditions, in fact under the exact same conditions required for strong duality in the risk-neutral setting. More specifically, we propose a general two-step compositional risk-constrained learning framework, where the risk incurred by each corresponding random loss (for each objective/constrained) is evaluated \textit{hierarchically}, \textit{first} relative to the response (e.g., label) posterior given the features, and \textit{then} relative to the feature prior. Essentially, we advocate for a \textit{decomposition of risk} into posterior risk and prior risk, where each component may be evaluated by a risk measure of different type.


In light of the above, our results pave the way for a more widespread use of risk measures in the context of constrained learning, and also for the development of dual-domain algorithms for actually solving such risk-aware constrained learning problems. 
At the same time, by specializing our results to the risk-neutral case, we resolve some technical issues present in the available literature (as discussed above), while also providing state-of-the-art results using more general and realistic assumptions, indeed verifiable in a practical setting. 

\paragraph*{\textbf{Notation}} Bold capital letters (such as ${\bf A}$), or calligraphic
letters (such as ${\cal A}$) will denote finite-dimensional sets/spaces, such as
Euclidean spaces. Math script letters (such as $\mathscr{A}$) will denote $\sigma$-algebras.
Boldsymbol letters (such as $\boldsymbol{A}$ or $\boldsymbol{a}$)
will denote (random) vectors. The space of $p$-integrable functions
from a measurable space $(\Omega,\mathscr{F})$ equipped with a ($\sigma$-)finite
measure $\mu \colon \mathscr{F}\rightarrow\mathbb{R}_{+}$ to a Banach space
$\mathbb{A}$, with standard notation ${\cal L}_{p}(\Omega,\mathscr{F},\mu;\mathbb{A})$, is abbreviated as ${\cal L}_{p}(\mu,\mathbb{A})$. The rest of the
notation is standard.

\section{Strong Duality in Risk-Constrained Functional Programming}\label{II}
Let us fix an arbitrary complete probability space $(\Omega,\mathscr{F},\mu)$,
and consider a random element $\boldsymbol{H}:\Omega\rightarrow{\cal H}\triangleq\mathbb{R}^{N_{\boldsymbol{H}}}$
with induced Borel measure $\mathrm{P}\equiv \mathrm{P}_{\boldsymbol{H}}:\mathscr{B}({\cal H})\rightarrow[0,1]$,
modeling some observable random phenomenon. Following the recent developments in \cite{KalogPougk:arXiv}, we are interested in
the class of risk-constrained nonconvex functional programs formulated as
\begin{equation}
\boxed{\begin{array}{rl}
\underset{\boldsymbol{x},\boldsymbol{p}(\cdot)}{\mathrm{maximize}} & g^{o}(\boldsymbol{x})\\
\mathrm{subject\,to} & \boldsymbol{x}\le-\boldsymbol{\rho}(-\boldsymbol{f}(\boldsymbol{p}(\boldsymbol{H}),\boldsymbol{H}))\\
 & \boldsymbol{g}(\boldsymbol{x})\ge{\bf 0}\\
 & (\boldsymbol{x},\boldsymbol{p})\in{\cal X}\times\Pi
\end{array},}\tag{{RCP}}\label{eq:Base}
\end{equation}
where  ${\bf C}\triangleq\mathbb{R}^{N}$, $g^{o}:\mathbb{R}^N\rightarrow\mathbb{R}$
and $\boldsymbol{g}:\mathbb{R}^N\rightarrow\mathbb{R}^{N_{\boldsymbol{g}}}$
are given concave functions, $\boldsymbol{p}:{\cal H}\rightarrow{\bf R}\triangleq\mathbb{R}^{N_{\boldsymbol{p}}}$
is an allocation policy on observables $\boldsymbol{H}$, $\boldsymbol{f}:{\bf R}\times{\cal H}\rightarrow{\bf C}$
is a (generally nonconvex) function measuring the quality of a policy $\boldsymbol{p}$ at each
realization $\boldsymbol{H}$ in ${\cal H}$ and such that $\boldsymbol{f}(\boldsymbol{p}(\cdot),\cdot)\in{\cal L}_{1}(\mathrm{P},{\bf C})$
on $\Pi$, and where $\boldsymbol{\rho}:{\cal L}_{1}(\mathrm{P},{\bf C})\rightarrow{\bf C}$
is a finite-valued vector risk measure, assumed to be \textit{convex,
lower semicontinuous and positively homogeneous} in every dimension
(i.e., component-wise), such that 
for
each $i\in\mathbb{N}_{N}^{+}$,
\[
\rho_{i}(\boldsymbol{Z})=\rho_{i}(Z_{i}),\quad\text{for all }\boldsymbol{Z}\in{\cal L}_{1}(\mathrm{P},{\bf C}).
\]
As indicated in \eqref{eq:Base}, the finite-dimensional variables $\boldsymbol{x}$ are further restricted to the set ${\cal X}\subseteq \mathbb{R}^N$ and the policies are restricted to some infinite-dimensional set $\Pi$.
\par It was recently proven in \cite{KalogPougk:arXiv} that \eqref{eq:Base} exhibits strong duality under rather standard assumptions, strictly generalizing state-of-the-art results on the risk-neutral case (i.e., where all associated risk measures are expectations), e.g., in \cite{Luo2008,Ribeiro2010,Ribeiro2012}.

\noindent\fcolorbox{black}{lightgray}{\begin{minipage}[t]{1\columnwidth - 2\fboxsep - 2\fboxrule}%
\begin{assumption}
\label{assu:Assumption}The following conditions are in effect:
\begin{enumerate}
\item The utilities $g^{o}$ and $\boldsymbol{g}$ are concave. 
\item The service feasible set ${\cal X}$ is convex.
\item The policy feasible set $\Pi$ is decomposable.
\item The Borel measure $\mathrm{P}$ is nonatomic\footnote{Recall that $\mathrm{P}$ is nonatomic if for any event $E$ with
$\mathrm{P}(E)>0$, an event $E'\subseteq E$ exists such that $\mathrm{P}(E)>\mathrm{P}(E')>0$.}.
\item Problem \textnormal{\eqref{eq:Base}} satisfies Slater's CQ (i.e., it is
strictly feasible).
\end{enumerate}
\end{assumption}
\end{minipage}}\vspace{8.5bp}
\noindent In particular, the following fact was established in \cite{KalogPougk:arXiv}.\\
\noindent\fcolorbox{black}{orange}{\begin{minipage}[t]{1\columnwidth - 2\fboxsep - 2\fboxrule}%
\begin{theorem}[Kalogerias, Pougkakiotis]
 \label{thm:Main} Let Assumption \textnormal{\ref{assu:Assumption}} be in effect.
Then problem \textnormal{\eqref{eq:Base}} has zero duality gap, i.e., $\mathsf{P}^{*}=\mathsf{D}^{*}$.
In fact, \textnormal{\eqref{eq:Base}} exhibits strong duality, i.e., optimal dual
variables exist.
\end{theorem}
\end{minipage}} \vspace{2.5bp}
\par In fact, Theorem \ref{thm:Main} is applicable in seemingly more general formulations of \eqref{eq:Base}. In particular, by letting $\rho_0$ be some real-valued convex, lower semicontinuous and positively homogeneous risk measure, $f_0  \colon \bm{R} \times \mathcal{H} \rightarrow \mathbb{R}$ be an arbitrary function such that $f_0(\bm{p}(\cdot),\cdot) \in \mathcal{L}_1(P,\mathbb{R})$, and $\bm{r} \colon \mathcal{X} \rightarrow \mathbb{R}^N$ be some component-wise convex function, it readily follows that the problem 
\begin{equation*}
\begin{array}{rl}
\underset{\boldsymbol{x},\boldsymbol{p}(\cdot)}{\mathrm{maximize}} & g^{o}(\boldsymbol{x}) - \rho_0(-f_0(\bm{p}(\bm{H}),\bm{H}))\\
\mathrm{subject\,to} & \bm{r}(\boldsymbol{x})\le-\boldsymbol{\rho}(-\boldsymbol{f}(\boldsymbol{p}(\boldsymbol{H}),\boldsymbol{H}))\\
 & \boldsymbol{g}(\boldsymbol{x})\ge{\bf 0}\\
 & (\boldsymbol{x},\boldsymbol{p})\in{\cal X}\times\Pi
\end{array},
\end{equation*}
\noindent also exhibits strong duality. For a detailed derivation of this fact, we refer the reader to \cite[Section 6.1]{KalogPougk:arXiv}.
\section{Nonconvex Risk-Constrained Learning}
\par In this section, we will provide precise conditions to show that a wide range of risk-constrained learning problems, involving nonconvex losses, can be cast in the form of \eqref{eq:Base}, thus exhibiting strong duality under Assumption \ref{assu:Assumption}.

\par We consider a general formulation of a supervised risk-constrained learning problem, where the associated loss functions are allowed to be nonconvex. On our usual probability space $(\Omega,\mathscr{F},\mu)$, we consider random \textit{example vectors} $(\boldsymbol{X}_i,Y_i) \colon \Omega \rightarrow \mathbb{R}^d \times \mathbb{R}$, $i \in \mathbb{N}_m$ together with their induced Borel probability distributions $\mathcal{D}_i:\mathscr{B}(\mathbb{R}^d \times \mathbb{R}) \rightarrow [0,1]$, instantiated over data pairs $(\boldsymbol{x},y)$, where $\boldsymbol{x} \in \mathbb{R}^d$ represents a realized feature or system input and $y \in \mathbb{R}$ represents a realized label or measurement (response).
We denote by $\mathcal{D}_{\boldsymbol{X}_i}$ the marginal probability distribution of $\boldsymbol{X}_i$, and by $\mathcal{D}_{Y_i\vert \boldsymbol{X}_i}$ the conditional probability distribution of $Y_i$ given a realization of $\boldsymbol{X}_i$, for $i \in \mathbb{N}_m$. 

\noindent\fcolorbox{black}{lightgray}{\begin{minipage}[t]{1\columnwidth - 2\fboxsep - 2\fboxrule}%
\begin{assumption} \label{assumption: nonatomicity}
The marginal probability distributions $\mathcal{D}_{\boldsymbol{X}_i}$, $i \in \mathbb{N}_m^+$, are absolutely continuous with respect to the ``common denominator" $\mathcal{D}_{\boldsymbol{X}_0}$ (without loss of generality), which in turn is assumed to be nonatomic.
\end{assumption}
\end{minipage}} 
\begin{remark}
We note that Assumption \ref{assumption: nonatomicity} implies, by virtue of the Radon-Nikodym theorem, that there must exist functions $w_i \colon \mathbb{R}^d \rightarrow \mathbb{R}^+$, such that $w_i \triangleq d\mathcal{D}_{\boldsymbol{X}_i}/d\mathcal{D}_{\boldsymbol{X}_0}$, for all $i \in \mathbb{N}_m^+$. Our assumption of absolute continuity of the marginal distributions $\mathcal{D}_{\boldsymbol{X}_i}$ with respect to $\mathcal{D}_{\boldsymbol{X}_0}$ is standard in the literature. Indeed, it holds if each distribution $\mathcal{D}_{\boldsymbol{X}_i}$ is assumed to have a Lebesgue density (e.g. see \cite{Chamon2021} and \cite{KalogPougk:arXiv}).  On the other hand, nonatomicity of the common denominator $\mathcal{D}_{\boldsymbol{X}_0}$ is also standard (e.g. see \cite[Theorem 1]{Chamon2021}) and very intuitive, since the feature or input space is naturally expected to be continuous in many applications of interest yielding (constrained) classification or constrained regression tasks.
\end{remark}
\par Before proceeding with our proposed formulation of a general risk-constrained learning problem, we introduce the notion of a \emph{conditional risk mapping}. In particular, for any $i \in \mathbb{N}_{m}$, we consider the vector spaces $\mathcal{Z}^i_{1} \triangleq \mathcal{L}_p(\Omega, \sigma(\bm{X}_i),\mu; \mathbb{R})$ and $\mathcal{Z}^i_{2} \triangleq \mathcal{L}_{p'}(\mathbb{R}^d\times \mathbb{R},\mathscr{B}(\mathbb{R}^d \times \mathbb{R}),\mathcal{D}_i) \equiv \mathcal{L}_{p'}(\mathcal{D}_i,\mathbb{R})$, where $p, p' \in [1,\infty]$. We define a conditional risk mapping as a functional $\widehat{\rho}\left(\cdot \big\vert \bm{X}_i\right) \colon \mathcal{Z}^i_{2} \rightarrow \mathcal{Z}^i_{1}$, with the property that for every qualifying Borel function $Z \colon \mathbb{R}^{d+1} \rightarrow \mathbb{R}$, $\widehat{\rho}(\cdot \vert \bm{X}_i)$ obeys the relation
\[ \widehat{\rho}\left(Z(\bm{X}_i,Y_i) \vert \bm{X}_i \right)(\omega) = \widehat{\rho}\left(Z(\bm{x},Y_i)\vert\bm{X}_i\right)(\omega)\big\vert_{\bm{x} =\bm{X}_i(\omega)},\]
\noindent for every elementary event $\omega \in \Omega$, where the instantiation $\left[\widehat{\rho}\left(Z(\bm{x},\cdot)\vert \bm{X}_i\right)\right](\omega) \colon \mathcal{Z}_3^i \rightarrow \mathbb{R}$ is a properly chosen risk functional, with $\mathcal{Z}^i_{3} \triangleq \mathcal{L}_{p'}(\mathbb{R},\mathscr{B}(\mathbb{R}),\mathcal{D}_{Y_i\vert \bm{X}_i}) \equiv \mathcal{L}_{p}(\mathcal{D}_{Y_i\vert \bm{X}_i},\mathbb{R})$.
A typical example of such a conditional risk mapping is, of course, the conditional expectation. We note that the definition of a conditional risk mapping above is compatible with our particular construction in Section \ref{II}, relying on a Borel space setting. Conditional risk mappings are usually defined more generally, and are often required to satisfy certain conditions of interest, and thus associated definitions and assumptions vary (e.g., see \cite[Definition 2.1]{Shapiro_Conditional_Risk_measures}, and \cite[Definition 1]{RIEDEL2004185}). Concrete examples of conditional risk mappings obeying our definition will be given later on.
\par Letting $\ell_i \colon \mathbb{R}^k \times \mathbb{R} \rightarrow \mathbb{R}_+$ be some given (possibly nonconvex or discontinuous) loss functions, for $i \in \mathbb{N}_m$, our proposed risk-constrained learning framework is based on the constrained nonconvex functional program
\begin{equation}
\hspace{-9pt}\boxed{\hspace{-5pt}\begin{array}{rl}
\:\underset{\boldsymbol{f}(\cdot) \in \mathcal{F}}{\mathrm{minimize}} & \hspace{-5pt}{\rho}_{0}\big(\widehat{{\rho}}_{0}\big( \ell_0\big(\boldsymbol{f}(\boldsymbol{X}_0),Y_0\big)\big\vert \boldsymbol{X}_0\big) \big) \\
\mathrm{subject\,to} & \hspace{-5pt}  {\rho}_{i}\big(\widehat{{\rho}}_{i}\big( \ell_i\big(\boldsymbol{f}(\boldsymbol{X}_i),Y_i\big)\big\vert \boldsymbol{X}_i\big) \big) \leq c_i, i \in \mathbb{N}_m^+
\end{array}\hspace{-4pt},} \hspace{-8pt}\tag{{RCL}}\label{eq:RCLearning}
\end{equation}
\noindent where $\boldsymbol{f} \colon \mathbb{R}^d \rightarrow \mathbb{R}^k$ belongs to some certain policy space $\mathcal{F}$, and for $i \in \mathbb{N}_m$, $c_i \in \mathbb{R}$, ${\rho}_{i}$ is a real-valued convex, lower semicontinuous, and positively homogeneous risk measure, while $\widehat{{\rho}}_{i}(\cdot\vert \boldsymbol{X}_i)$ is some conditional risk mapping.

\noindent\fcolorbox{black}{lightgray}{\begin{minipage}[t]{1\columnwidth - 2\fboxsep - 2\fboxrule}%
\begin{assumption}\label{assumption: conditional risk measure}
    For each $i \in \mathbb{N}_m$, we have $\ell_i\left(\bm{f}(\bullet),\cdot\right) \in \mathcal{Z}^i_{2}$, for any $\bm{f} \in \mathcal{F}$, and the space of policies $\mathcal{F}$ is decomposable.
\end{assumption}
\end{minipage}} 

\begin{remark}
Let us emphasize that Assumption \ref{assumption: conditional risk measure} is very general. The decomposability of the policy space $\mathcal{F}$ is standard, and is already utilized in \eqref{eq:Base}. Furthermore, the assumption that $\ell_i\left(\bm{f}(\bullet),\cdot\right) \in \mathcal{Z}^i_{2}$ is also standard and very mild. In particular, it also appears in Assumption \ref{assu:Assumption}, and includes various nonconvex or even vastly discontinuous functions $\ell_i$.
\end{remark}

    Fix $i \in \mathbb{N}_m$. Under Assumption \ref{assumption: conditional risk measure}, and using our definition of a conditional risk mapping $\widehat{\rho}(\cdot\vert\bm{X}_i)$, it readily follows that
    \begin{equation*}
   \begin{split}
    \widehat{{\rho}}_{i}( \ell_i(\boldsymbol{f}(\boldsymbol{X}_i),Y_i)\vert \boldsymbol{X}_i)  & \equiv \widehat{{\rho}}_{i}( \ell_i(\boldsymbol{z},Y_i)\vert \boldsymbol{X}_i)\big\vert_{\boldsymbol{z} = \boldsymbol{f}(\boldsymbol{X}_i)} \\ & \equiv F_i\left(\boldsymbol{f}(\boldsymbol{X}),\boldsymbol{X}\right),
    \end{split}
    \end{equation*}
    \noindent such that $F_i\left(\boldsymbol{f}(\cdot),\cdot\right) \in \mathcal{L}_p\big(\mathcal{D}_{\boldsymbol{X}_i},\mathbb{R}\big)$, for any $\boldsymbol{f} \in \mathcal{F}$. We now showcase that our definition of conditional risk mappings is general and includes various useful examples as special cases, as follows:
\begin{itemize}
    \item As already mentioned, the most typical example of a conditional risk mapping is that of conditional expectation. In the context of \eqref{eq:RCLearning}, we set $\widehat{\rho}(\cdot \big\vert \bm{X}_i) = \mathbb{E}\left\{\cdot \big\vert \bm{X}_i\right\}.$ Since we assume that the conditional distribution of $\bm{Y}_i$ given $\bm{X}_i$ exists, it then readily follows that 
    \[ \mathbb{E}\left\{\ell_i\left(\bm{f}(\bm{X}_i),Y_i\right) \vert \bm{X}_i\right\} = \mathbb{E}\left\{\ell_i\left(\bm{z},Y_i\right) \big\vert \bm{X}_i\right\}\big\vert_{\bm{z} = \bm{f}(\bm{X}_i)}.\]
    \noindent The latter equality is well-known as the \emph{substitution rule}.
    \item Let us now consider two popular cases in the class of \textit{coherent} conditional risk mappings (see \cite[Chapter 6]{Shapiro_Stochastic_Prog}). The first is the conditional version of the conditional value at risk (CVaR). For any $i \in \mathbb{N}_m$, and any $\alpha \in (0,1)$, we can define $\textnormal{CVaR}^i_{\alpha}\left(\cdot \vert \bm{X}_i\right) \colon \mathcal{Z}^i_{2} \rightarrow \mathcal{Z}^i_{1}$, as
    \[ \textnormal{CVaR}^i_{\alpha}\left(Z \vert \bm{X}_i\right) \hspace{-2pt} = \hspace{-2pt} \inf_{W \in \mathcal{Z}^i_{1}}\left\{W + \alpha^{-1} \mathbb{E}\left\{\left(Z - W\right)_+ \vert \bm{X}_i\right\}\right\},\]
    \noindent for $Z \in \mathcal{Z}_2^i$, where $(\cdot)_+ \equiv \max\{\cdot,0\}$. As before, the instantiation $\left[\textnormal{CVaR}^i_{\alpha}\left(Z \vert \bm{X}_i\right)\right](\bm{X}(\omega))$ can be considered to take values from $\mathcal{Z}_3^i$. As in the case of conditional expectation, it is not hard to see that by letting $Z = \ell_i(\bm{f}(\bm{X}_i),Y_i)$, we obtain
    \begin{equation*}
\begin{split}
& \textnormal{CVaR}^i_{\alpha}\left(\ell_i(\bm{f}(\bm{X}_i),Y_i) \vert \bm{X}_i\right) \\ 
& \qquad = \textnormal{CVaR}^i_{\alpha}\left(\ell_i(\bm{z},Y_i) \vert \bm{X}_i\right)\big\vert_{\bm{z} = \bm{f}(\bm{X}_i)}.
    \end{split}
    \end{equation*}
    \noindent The second popular case we consider is the conditional mean-upper-semideviation. In particular, for $Z \in \mathcal{Z}_{2}^i$, and some $c \in [0,1]$, this conditional risk mapping takes the form
    \begin{equation*}
\begin{split}
 \widehat{\rho}(Z\vert \bm{X}_i) & \hspace{-2pt}=\hspace{-2pt} \mathbb{E}\left\{Z \vert \bm{X}_i\right\} \hspace{-2pt}+\hspace{-2pt} c \big(\mathbb{E}\big\{ \left(Z \hspace{-2pt}-\hspace{-2pt} \mathbb{E}\{Z \vert \bm{X}_i \}\right)^{p'}_+\big\vert \bm{X}_i\big\}\big)^{1/p'}.
\end{split}
\end{equation*}
\noindent Again, one can readily verify that this conditional risk mapping is well-defined in the sense of our definition, and thus satisfies the substitution rule.
\item Lastly, in order to stress the generality of the conditional risk mappings considered in this work, let us define conditional extensions of the generalized mean semideviations introduced in \cite{Kalogerias2018b}. In fact, we may consider a larger class of generalized mean semideviations than those discussed in \cite{Kalogerias2018b}. To that end, let $R \colon \mathbb{R} \rightarrow \mathbb{R}$ be any nonnegative (possibly nonconvex) function. For any $Z \in \mathcal{Z}_{2}^i$, and any $c \in [0,+\infty)$, we consider conditional generalized mean semideviation risk measures, defined as
\begin{equation*}
\begin{split} &\widehat{\rho}(Z\vert \bm{X}_i) \\
&= \mathbb{E}\left\{Z \vert \bm{X}_i\right\} + c \big(\mathbb{E}\big\{ \left(R\left(Z - \mathbb{E}\{Z \vert \bm{X}_i \}\right)\right)^{p'}\big\vert \bm{X}_i\big\} \big)^{1/p'}, 
\end{split}
\end{equation*}
\noindent provided that $R(Z - \mathbb{E}\{Z \vert \bm{X}_i \}) \in \mathcal{Z}_{1}^i$. Apparently, this conditional risk mapping is also well-defined according to our definition and satisfies the substitution rule. At the same time, we observe that, depending on the choice of $R$, such conditional risk mappings might fail to satisfy several properties such as convexity, monotonicity, positive homogeneity, etc. The model in \eqref{eq:RCLearning} allows such general constructions, showcasing the wide range of conditional risk mappings enabled in our framework.
\end{itemize}

Our first main result in this section may now be formulated, as follows. 

\noindent\fcolorbox{black}{orange}{\begin{minipage}[t]{1\columnwidth - 2\fboxsep - 2\fboxrule}%
\begin{theorem} \label{thm: problem reformulation}
    Let problem \eqref{eq:RCLearning} satisfy Assumptions \textnormal{\ref{assumption: nonatomicity}} and \textnormal{\ref{assumption: conditional risk measure}}. Then, there exist convex, proper, lower-semicontinuous and positively homogeneous risk measures $\widetilde{\rho_i} \colon \mathcal{L}_1(\mathcal{D}_{\bm{X}_0},\mathbb{R}) \rightarrow \mathbb{R}$, as well as functions $G_i(\bullet,\cdot)$ satisfying $G_i(\bm{f}(\cdot),\cdot) \in \mathcal{L}_1(\mathcal{D}_{\bm{X}_0},\mathbb{R})$, for all $i \in \mathbb{N}_m^+$, such  that problem \eqref{eq:Base} can be equivalently written as
    \begin{equation}
\hspace{-4pt}\begin{array}{rl}
\:\underset{\boldsymbol{f}(\cdot) \in \mathcal{F}}{\mathrm{minimize}} &{\rho}_{0}\left(F_0\big(\boldsymbol{f}(\boldsymbol{X}_0),\boldsymbol{X}_0\big) \right) \\
\mathrm{subject\,to} &  \widetilde{{\rho}}_{i}\left( G_i\big(\boldsymbol{f}(\boldsymbol{X}_0),\boldsymbol{X}_0\big)\right) \leq c_i, i \in \mathbb{N}_m^+
\end{array}.\tag{{RCL0}}\label{eq:RCLearning0}
\end{equation}
\end{theorem}
\end{minipage}} 
\begin{proof}

\par From Assumption \ref{assumption: nonatomicity}, we observe that each $\mathcal{D}_{\boldsymbol{X}_i}$, for $i \in \mathbb{N}_m^+$, is absolutely continuous with respect to $\mathcal{D}_{\boldsymbol{X}_0}$, while, from Assumption \ref{assumption: conditional risk measure}, we have shown that $F_i(\boldsymbol{f}(\cdot),\cdot) \in \mathcal{L}_p(\mathcal{D}_{\boldsymbol{X}_i},\mathbb{R}) \subset \mathcal{L}_1(\mathcal{D}_{\boldsymbol{X}_i},\mathbb{R})$. It then follows that $G_i(\boldsymbol{f}(\cdot),\cdot) \triangleq F_i(\boldsymbol{f}(\cdot),\cdot) w_i \in \mathcal{L}_1(\mathcal{D}_{\boldsymbol{X}_0},\mathbb{R})$, $i \in \mathbb{N}_m^+$. Let $\mathbb{A}_{i} \subseteq \mathcal{L}_{\infty}(\mathcal{D}_{\boldsymbol{X}_i},\mathbb{R})$, for all $i \in \mathbb{N}_m$, be the uncertainty sets corresponding to $\rho_i(\cdot)$ and consider another (related) convex, lower-semicontinuous, and positively homogeneous risk measure $\widetilde{\rho}_i(\cdot)$ defined on $w_i \mathcal{L}_1(\mathcal{D}_{\boldsymbol{X}_i},\mathbb{R})\subseteq \mathcal{L}_1(\mathcal{D}_{\boldsymbol{X}_0},\mathbb{R})$, such that $\widetilde{\rho}_i(Zw_i) = \rho(Z)$, for any $Z \in \mathcal{L}_1(\mathcal{D}_{\boldsymbol{X}_i},\mathbb{R})$. In particular, we define 
\begin{equation*}
\begin{split}
\widetilde{\rho}_{i}\left( G_i\big(\boldsymbol{f}(\boldsymbol{X}_0),\boldsymbol{X}_0\big)\right) \triangleq &\ \sup_{\zeta \in \mathbb{A}_i} \int \zeta(\boldsymbol{x})  G_i\big(\boldsymbol{f}(\boldsymbol{x}),\boldsymbol{x} \big)d\mathcal{D}_{\boldsymbol{X}_0}(\boldsymbol{x}) \\  = &\ \sup_{\zeta \in \mathbb{A}_i} \int \zeta(\boldsymbol{x})  F_i\big(\boldsymbol{f}(\boldsymbol{x}),\boldsymbol{x} \big)d\mathcal{D}_{\boldsymbol{X}_i}(\boldsymbol{x}) \\
=&\ {\rho}_{i}\left( F_i\big(\boldsymbol{f}(\boldsymbol{X}_i),\boldsymbol{X}_i\big)\right) \\ =&\ {\rho}_{i}\big(\widehat{{\rho}}_{i}\big( \ell_i\big(\boldsymbol{f}(\boldsymbol{X}_i),Y_i\big)\Big \vert \boldsymbol{X}_i\big) \big),
\end{split}
\end{equation*}
\noindent where we have used the fact that $\rho_i(\cdot)$ is convex, lower-semicontinuous, and positively homogeneous. However, the supremum  in the second integral, defining ${\rho}_i(\cdot)$, would not change by restricting $\mathbb{A}_i$ to an appropriately selected bounded set $\widetilde{\mathbb{A}}_i \subseteq \mathcal{L}_{\infty}(\mathcal{D}_{\bm{X}_0},\mathbb{R})$ chosen independently of $Z w_i$ and with $\mathbb{A}^i \supseteq \widetilde{\mathbb{A}}^i$, since the integral is taken with respect to $d\mathcal{D}_{\bm{X}_i}$, and does not change for functions differing on $\mathcal{D}_{\bm{X}_i}$-measure zero sets. In other words, $\widetilde{\rho}(\cdot)$ admits a representation
\[ \widetilde{\rho}_{i}\left( G_i\big(\boldsymbol{f}(\boldsymbol{X}_0),\boldsymbol{X}_0\big)\right) 
 \equiv \sup_{\zeta \in \widetilde{\mathbb{A}}_i} \int \zeta(\boldsymbol{x})  G_i\big(\boldsymbol{f}(\boldsymbol{x}),\boldsymbol{x} \big)d\mathcal{D}_{\boldsymbol{X}_0}(\boldsymbol{x}),  \]
 \noindent for some $\widetilde{\mathbb{A}}_i \subseteq \mathcal{L}_{\infty}(\mathcal{D}_{\boldsymbol{X}_0},\mathbb{R})$. Using the introduced notation, we recast \eqref{eq:RCLearning} into the equivalent (given our assumptions) form of \eqref{eq:RCLearning0}. 
 \end{proof}
Leveraging Theorem \ref{thm: problem reformulation}, we can now state our second main result of this section.

 \noindent\fcolorbox{black}{orange}{\begin{minipage}[t]{1\columnwidth - 2\fboxsep - 2\fboxrule}%
\begin{theorem} \label{thm: final result}
    Let Assumptions \textnormal{\ref{assumption: nonatomicity}}, \textnormal{\ref{assumption: conditional risk measure}} hold for problem \textnormal{\eqref{eq:RCLearning}}. If, additionally, \textnormal{\eqref{eq:RCLearning}} satisfies Slater's constraint qualification, it exhibits strong duality.
\end{theorem}
\end{minipage}} 
\begin{proof}
\noindent We note that under Assumptions \ref{assumption: nonatomicity}, \ref{assumption: conditional risk measure}, and by utilizing Theorem \ref{thm: problem reformulation}, \eqref{eq:RCLearning0} is an instance of \eqref{eq:Base}, and hence Staler's constraint qualification suffices to ensure that it satisfies the conditions given in Assumption \ref{assu:Assumption}. It then follows from Theorem \ref{thm:Main} that \eqref{eq:RCLearning0} exhibits strong duality, and of course the same holds for \eqref{eq:RCLearning}.
\end{proof}
\section{The risk-neutral case}
\par So far, we have stated problem \eqref{eq:RCLearning} (and consequently \eqref{eq:RCLearning0}) in full generality. A highly important instance of \eqref{eq:RCLearning} is in the risk-neutral setting, where $\rho_i(\cdot) = \mathbb{E}_{\mathcal{D}_{\boldsymbol{X}_i}}\{\cdot\}$, and $\widehat{\rho}_i(\cdot\vert \boldsymbol{X}_i) = \mathbb{E}\{\cdot \vert \bm{X}_i\}$, for all $i \in \mathbb{N}_m$. Using the tower property, it is easy to see that, for all $i \in \mathbb{N}_m$,
\begin{equation*}
\begin{split}
&{\mathbb{E}}_{\mathcal{D}_{\bm{X}_i}}\left\{ \mathbb{E}\Big\{ \ell_i\big(\boldsymbol{f}(\boldsymbol{X}_i),Y_i\big)\big\vert \bm{X}_i\Big\} \right\} = \mathbb{E}_{\mathcal{D}_{i}} \left\{  \ell_i\big(\boldsymbol{f}(\boldsymbol{X}_i),Y_i\big) \right\},
\end{split}
\end{equation*}
\noindent where $\mathcal{D}_{i}$ is the Borel probability distribution of $(\bm{X}_i,Y_i)$. Thus, the risk-neutral constrained learning problem reads
\begin{equation}
\boxed{\begin{array}{rl}
\:\underset{\boldsymbol{f}(\cdot) \in \mathcal{F}}{\mathrm{minimize}} &{\mathbb{E}}_{\mathcal{D}_{0}}\left\{ \ell_0\big(\boldsymbol{f}(\boldsymbol{X}_0),Y_0\big)\right\} \\
\mathrm{subject\,to} &  {\mathbb{E}}_{\mathcal{D}_{i}}\left\{ \ell_i\big(\boldsymbol{f}(\boldsymbol{X}_i),Y_i\big) \right\} \leq c_i, \quad i \in \mathbb{N}_m^+
\end{array}.}\tag{{CL}}\label{eq:risk-neutral CLearning}
\end{equation}
Variations of this problem have been heavily studied and utilized in the machine learning  literature (e.g. \cite{JOTA_Shapiro,primal_dual_constrained_learning,NEURIPS2021_Chamon,Chamon2021}). By considering \eqref{eq:risk-neutral CLearning}, let us assume that $\mathcal{D}_{\boldsymbol{X}_0}$ is nonatomic (without loss of generality), and that $\mathcal{D}_{\boldsymbol{X}_0} \gg \mathcal{D}_{\boldsymbol{X}_i}$, that Slater's CQ holds, and that $\ell_i(\bm{f}(\bullet),\cdot) \in \mathcal{L}_1(\mathcal{D}_{i},\mathbb{R})$ for any $\boldsymbol{f} \in \mathcal{F}$, where $\mathcal{F}$ is a decomposable functional space. Under this very general framework, we claim that the problem exhibits strong duality. In fact, we recover the results of \cite[Proposition III.2 and Proposition B.1]{Chamon2021} (unifying the classification and regression regimes), while dispensing several additional assumptions.
\par Before we compare our result to that given in \cite{Chamon2021}, let us first discuss its proof. Obviously, we could directly apply Theorem \ref{thm: final result} (based on Theorem \ref{thm:Main}, the proof of which can be found in \cite[Section 5]{KalogPougk:arXiv}). While this would certainly be a possibility, the analysis in \cite{KalogPougk:arXiv} utilizes Uhl's weak extension of A. A. Lyapunov convexity theorem (see \cite[Theorem 1]{Uhl1969}), since the latter is applicable to a wide range of nonlinear functionals (and strictly more general than expectations). While this is a very general result, it requires that the underlying nonatomic measure is finite (thus not directly covering $\sigma$-finite measures). Nonetheless, using the developments of the previous section, we observe that problem \eqref{eq:risk-neutral CLearning} can be equivalently written as
\begin{equation}
\begin{array}{rl}
\:\underset{\boldsymbol{f}(\cdot) \in \mathcal{F}}{\mathrm{minimize}} &{\mathbb{E}}_{\mathcal{D}_{0}}\left\{ F_0\big(\boldsymbol{f}(\boldsymbol{X}_0),\bm{X}_0\big)\right\} \\
\mathrm{subject\,to} &  {\mathbb{E}}_{\mathcal{D}_{0}}\left\{ G_i\big(\boldsymbol{f}(\boldsymbol{X}_0),\bm{X}_0\big) \right\} \leq c_i, i \in \mathbb{N}_m^+
\end{array},\tag{{CL0}}\label{eq:risk-neutral CLearning reformulated}
\end{equation}
\noindent where $F_0(\bm{f}(\cdot),\cdot),\ G_i(\bm{f}(\cdot),\cdot) \in \mathcal{L}_1(\mathcal{D}_{\bm{X}_0},\mathbb{R})$, for all $i \in \mathbb{N}_m^+.$ In then follows that a simple application of the standard A. A. Lyapunov's convexity theorem (see \cite[Corollary IX.1.6]{Diestel1977}), which also supports $\sigma$-finite nonatomic measures (such as the Lebesgue measure), combined with an appropriate application of the supporting hyperplane theorem (e.g., see \cite[Proposition 1.5.1]{Bertsekas2009}), would immediately yield the desired result, that is \eqref{eq:risk-neutral CLearning reformulated} (and thus \eqref{eq:risk-neutral CLearning}) exhibits strong duality. Indeed, since we have already shown the equivalence between \eqref{eq:risk-neutral CLearning} and \eqref{eq:risk-neutral CLearning reformulated}, the strong duality result follows immediately by \cite[Theorem 1]{Ribeiro2012} (see also prior developments in \cite{Luo2008,Ribeiro2010}).
\par Let us now compare our (risk-neutral) strong duality result with that shown in \cite{Chamon2021}. The major difference lies in the way that the authors in \cite{Chamon2021} condition the problem. In particular, \cite{Chamon2021} relies on the nonatomicity of conditional distributions of the form $\mathcal{D}_{\bm{X}_i\vert Y_i}$, in contrast to our work, that requires nonatomicity of the marginal distributions $\mathcal{D}_{\bm{X}_i}$. We argue that our approach has significant benefits. Firstly, the model studied in \cite{Chamon2021} assumes that $\mathcal{F}$ is closed, convex, and decomposable. Instead, we have shown that only decomposability is required for showing strong duality of \eqref{eq:risk-neutral CLearning}. Secondly, our result is applicable to both classification and regression tasks, under the same set of (minimal) assumptions. In contrast, the aforementioned work separates these two cases. Specifically, the strong duality result for regression problems given in \cite[Proposition B.1]{Chamon2021} is shown here to hold without the additional requirement postulated in \cite[Assumption 6]{Chamon2021} (which, in essence, requires uniform continuity of $\ell_i(\bm{f}(\bullet),y) w_i(\cdot\vert y)$, for all $\bm{f} \in \mathcal{F}$, where $w_i(\cdot\vert Y_i)$ is the density induced by $\mathcal{D}_{\bm{X}_i\vert Y_i}$; note that this condition might be extremely difficult to verify and is quite restrictive). In fact, \cite{Chamon2021} makes some further implicit assumptions, including boundedness of the range of ${Y}_i$, for all $i \in \mathbb{N}_m$, as well as Lipschitz continuity of $\ell_i(\bullet,y)$ for all $y$ in the range of ${Y}_i$ (although the latter is also required to show some results that are not related to strong duality).
\par Apart from the benefits of the proposed approach, we should also mention a technical issue appearing in the analysis given in \cite{Chamon2021}, while proposing a way to fix it. In particular, in the proof of \cite[Lemma B.2]{Chamon2021}, the last argument utilized by the authors to show that the corresponding ``cost-constrained set" (denoted as $\mathcal{C}$ in \cite{Chamon2021}) is convex, is not accurate (to the best of our knowledge). Indeed, what the authors may show at most is that the \textit{closure} of this set is convex, and thus their proof is incomplete. Nonetheless, this can be fixed by utilizing the developments in \cite[Section 5.3]{KalogPougk:arXiv}, where it is shown that convexity of the closure of the ``cost-constrained set" suffices to ensure strong duality of \eqref{eq:risk-neutral CLearning}. Of course, in our analysis, this step is completely bypassed.
\section{Conclusion}
\vspace{-3.5pt}
In this paper, we established strong duality for a wide class of risk-constrained learning problems. Our results rely on a recent result relying on an application of Uhl's extension of A. A. Lyapunov's convexity theorem for general, infinite dimensional Banach spaces in the context of infinite-dimensional optimization, and are applicable to problems involving a wide range of risk measures, strictly generalizing existing results available for the risk-neutral constrained learning setting, and without imposing additional assumptions. We proposed a general risk-constrained functional learning framework involving nonconvex (possibly even discontinuous) losses and two-step compositional risk measures. The outer risk measures (evaluating feature risk) are assumed to be real-valued, convex, lower semicontinuous and positively homogeneous, with support over the  space $\mathcal{L}_1$, while the inner (conditional) risk mappings (evaluating posterior risk) are significantly more general, and are even allowed to be nonconvex. In the special case of risk-neutral constrained learning, we unified existing results for constrained regression and constrained classification tasks, while dispensing several assumptions utilized in the current literature. Overall, we have presented new state-of-the-art strong duality relations for a rich risk-constrained learning framework, hopefully paving the way for a more widespread utilization of risk-constraints in this setting.

\bibliographystyle{IEEEbib}
\bibliography{references.bib}
\end{document}